\documentclass[12pt,onecolumn,journal]{IEEEtran}

\usepackage{epsfig}
\usepackage{epstopdf}
\usepackage{citesort}
\usepackage{amsmath}
\usepackage{amssymb}
\usepackage{color}
\usepackage{array}
\usepackage{multirow}
\usepackage{algorithm}
\usepackage{algorithmic}
\usepackage{rotating}
\usepackage{graphicx}
\usepackage{stywhispers}
\newlength{\figurewidth}
\newlength{\smallfigurewidth}

\begin{document}

\title
{
SrTR: Self-reasoning Transformer with Visual-linguistic Knowledge for Scene Graph Generation
}

\name{Yuxiang Zhang$^{1}$, Zhenbo Liu$^{2}$, Shuai Wang$^{3}$}

\address{$^1$School of Information and Electronics, Beijing Institute of Technology
	\\ $^2$  Noah's Ark Lab, Huawei \\ $^3$ Department of Chemistry, The University of Hong Kong
	\thanks{corresponding author:  zyx829625@163.com}}


\maketitle
\thispagestyle{empty}
\pagestyle{empty}


\begin{abstract}
Objects in a scene are not always related. The execution efficiency of the one-stage scene graph generation approaches are quite high, which infer the effective relation between entity pairs using sparse proposal sets and a few queries. However, they only focus on the relation between subject and object in triplet set {\sffamily \textless \textbf{s}ubject entity, \textbf{p}redicate entity, \textbf{o}bject entity\textgreater }, ignoring the relation between subject and predicate or predicate and object, and the model lacks self-reasoning ability. In addition, linguistic modality has been neglected in the one-stage method. It is necessary to mine linguistic modality knowledge to improve model reasoning ability. To address the above-mentioned shortcomings, a Self-reasoning Transformer with Visual-linguistic Knowledge (SrTR) is proposed to add flexible self-reasoning ability to the model. An encoder-decoder architecture is adopted in SrTR, and a self-reasoning decoder is developed to complete three inferences of the triplet set, \textbf{s+o→p}, \textbf{s+p→o} and \textbf{p+o→s}. Inspired by the large-scale pre-training image-text foundation models, visual-linguistic prior knowledge is introduced and a visual-linguistic alignment strategy is designed to project visual representations into semantic spaces with prior knowledge to aid relational reasoning. Experiments on the Visual Genome dataset demonstrate the superiority and fast inference ability of the proposed method.
\end{abstract}


\section{Introduction}
Scene graph generation (SGG) is an advanced semantic understanding task based on object detection, which aims to provide a graphical representation of objects and their relationships in images. Vision and natural language are connected by a scene graph, which provides a local semantic representation of the visual scene and is considered useful in many visual tasks, such as visual question answering \cite{JustinJohnson2017InferringAE,JiaxinShi2018ExplainableAE,VinayDamodaran2021UnderstandingTR}, image captioning \cite{JingZhang2020EmpoweringTW,XuYang2018AutoEncodingSG}, image generation \cite{JustinJohnson2018ImageGF,HelisaDhamo2020SemanticIM}, and image retrieval \cite{JustinJohnson2015ImageRU}.

The scene graph is composed of a triplet set {\sffamily \textless subject entity, predicate, object entity\textgreater }, in which the entity is the node and the predicate is the edge. The graph structure is used to model the scene and mine the possible relationships between all entities in the scene. The development of SGG has been divided into two stages and one stage from the method level. In the two-stage method, they use existing object detectors (e.g. FasterRCNN \cite{ShaoqingRen2015FasterRT}) to detect entity proposals, and then input the relational inference module to predict the predicate classes of these entity combinations. Although this strategy achieves high recalls in SGG tasks, their computational cost is quite high, because there is too much predicate proposals for the combination of all objects in the scene. Recently, one-stage object detectors have been widely concerned (e.g. Detection Transformer, DETR \cite{KirillovAlexander2020EndtoEndOD}). They regard object detection as an end-to-end set prediction task and propose set-based loss through bipartite matching. This prompted the development of the one-stage SGG. After obtaining the entity queries through DETR, a fixed number of relation queries are set and sent to the relation transformer for relational reasoning. In this strategy, there is no need to pair the objects to predict all relationships, which has great development potential due to its fast inference speed and low computational cost.

Most one-stage techniques use custom relation queries and entity queries extracted by Convolutional Neural Networks (CNN) backbone to learn relations via relation transformer, which has limited model self-reasoning ability and flexibility. Self-reasoning refers to the ability to automatically infer the third element from two of the known elements in the triple, such as the model can autonomously infer the corresponding possible object when it knows subject and predicate. Furthermore, linguistic modality is rarely taken into account in the one-stage approach. Text has been shown in multimodal learning to be helpful for visual representation learning, and linguistic modality can well depict the relationship between objects. As a result, it is critical to understand how to leverage prior information in linguistic modality to guide relation transformer in learning the relationship between objects.

In order to address the aforementioned problems, a straightforward multi-modal self-reasoning SGG framework, called  Self-reasoning Transformer with Visual-linguistic Knowledge (SrTR), is proposed. SrTR is made up of three major modules: entity decoder, self-reasoning decoder, and visual-linguistic alignment. First, we introduce CNN and Deformable DETR for extracting the encoded multi-scale visual feature contexts. To obtain the entity representation, the Deformable DETR decoder is adopted as entity decoder and utilized to interactively encode entity queries and visual features. In addition, to implement the self-reasoning training of the model, we construct a self-reasoning decoder. {\sffamily \textless \textbf{s}ubject entity, \textbf{p}redicate, \textbf{o}bject entity\textgreater } is defined in the form of triplet query, and three inferences of the triplet set, \textbf{s+o→p}, \textbf{s+p→o} and \textbf{p+o→s}, are completed in the self-reasoning decoder. Finally, the linguistic features of the triplet set are generated using the large-scale trained CLIP \cite{AlecRadford2021LearningTV}, and the visual-linguistic alignment is designed to introduce visual-linguistic prior knowledge. Using supervised contrastive learning, the visual representation of the triplet set output by the self-reasoning encoder is projected into a prior semantic space, which strengthens the representation learning and assists the relational reasoning.

The main contributions of this work are summarized as follows.

\begin{itemize}
	\item A relational self-reasoning module is proposed to learn the potential relationship among subject, object and predicate in the training phase.
	
	\item Integrating multi-scale features and bounding box into triplet query for embedding visual and location information needed in the self-reasoning process.
	
	\item Visual-linguistic prior information is introduced into the one-stage SGG framework, and visual-linguistic alignment is designed to the subject, object and predicate to aid the self-reasoning of model.
\end{itemize}

The rest of the paper is organized as follows. Section II introduces relevant concepts of SGG. Section III elaborates on the proposed SrTR. The extensive experiments and analyses are presented in Section IV. Finally, conclusions are drawn in Section V.

\section{Related Work}
\label{sec:related-works}


\subsection{Two-stage Scene Graph Generation}

Existing SGG works primarily concentrate on improving the context modeling structure \cite{DanfeiXu2017SceneGG,XinLin2022RUNetRU,ZhangJi2019GraphicalCL,TzuJuiJuliusWang2020TacklingTU,JianweiYang2018GraphRF} or solving the class imbalance problem \cite{AlakhDesai2021LearningOV,MohammedSuhail2021EnergyBasedLF,RongjieLi2021BipartiteGN,KaihuaTang2020UnbiasedSG,XingningDong2022StackedHA} (i.e., long-tail distribution). Wang et al. \cite{TzuJuiJuliusWang2020TacklingTU} addressed the issue of annotation bias and sparse annotation in visual genome (VG) and suggested a new SGG training scheme with two relation classifiers, one of which offers less biased settings for the other. Based on the graph neural network-based message passing, Lin et al. \cite{XinLin2022RUNetRU} proposed a Regularized Unrolling Network ( RU-Net ) to solve the problem that it is very sensitive to the correlation between spurious nodes. To lessen noise in context modeling, Li et al. \cite{RongjieLi2021BipartiteGN} employed a relationship prediction confidence-based adaptive message transmission technique. Tang et al. \cite{KaihuaTang2020UnbiasedSG} suggested an unbiased approach that eliminates the vision-agnostic bias with counterfactual causality. The above complex two-stage methods usually need to predict the relationship between densely connected entity pairs, resulting in high computational cost and poor representation learning flexibility, and also limits end-to-end optimization.

\subsection{One-stage Scene Graph Generation}

One-stage target detection due to its excellent performance, recent researches have begun to explore the one-stage SGG framework. These first-level methods use a fully convolutional network \cite{HengyueLiu2021FullyCS,YaoTeng2021StructuredSR} or Transformer \cite{YurenCong2022RelTRRT,RajatKoner2020RelationTN} architecture to detect relationships directly from visual features without pre-training the target detection, and are simple, fast, and easy to train. Liu et al. \cite{HengyueLiu2021FullyCS} proposed a fully convolutional scene graph generation (FCSGG), which is the first fully convolutional-based one-stage SGG framework. The object is encoded as the center point of the bounding box, and the relationship is encoded as a two-dimensional vector field. Cong et al. \cite{YurenCong2022RelTRRT} presented an end-to-end SGG model Relation Transformer (RelTR) inspired by the advantages of DETR in object detection. The approach is made up of the encoder-decoder structure. The encoder infers the visual feature context, whereas the decoder infers a set of fixed-size subject-predicate-object triples using various types of attention mechanisms with coupled subject and object queries.

\section{Proposed Self-reasoning Transformer}
\label{sec:proposed}

\begin{figure*}[tp] \small
	\vspace{-2em}
	\begin{center}
		\centering
		\epsfig{width=2.2\figurewidth,file=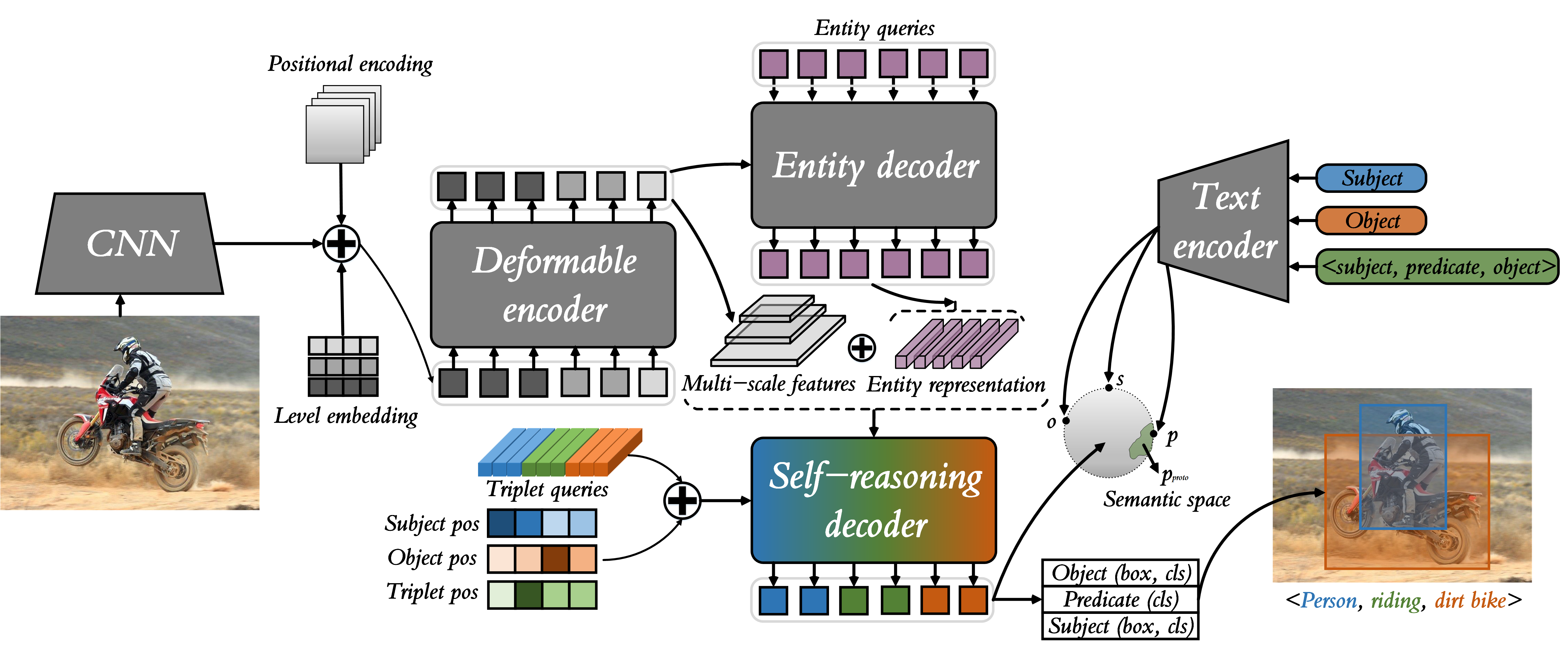}
		\caption{\label{fig:Flowchart}
			Flowchart of the proposed SrTR. One image is encoded by CNN, Deformable DETR encoder and decoder to obtain multi-scale visual features and entity representation. The predicted bounding box and multi-scale visual features complete the initialization of the triplet queries. The decoupled triplet queries are input into the self-reasoning decoder for self-reasoning training. The linguistic features of the triplet set are obtained by using the CLIP encoder, and visual-linguistic priors are introduced to make visual-linguistic alignment with the visual embedding features obtained by the self-reasoning encoder.}
	\end{center}
	\vspace{-2em}
\end{figure*}

The proposed SrTR is broken into three parts: an entity decoder, a self-reasoning decoder, and a visual-linguistic alignment. The flowchart for each part is illustrated in Fig.~\ref{fig:Flowchart}. (1) \textbf{Backbone and Entity Decoder}: Given an image, a multi-scale visual feature context ${\bf{M}}_e$ is encoded by a CNN and Deformable DETR encoder, which is then delivered to the Deformable DETR decoder, which interacts with entity query ${\bf{Q}}_e$ to produce the entity representation ${\bf{H}}_e$ and its matching bounding box ${\bf{B}}_e$. (2) \textbf{Self-reasoning Decoder}: In order to obtain multi-scale entity visual features ${\bf{Fea}}_e$, we map the bounding box ${\bf{B}}_e$ back to multi-scale space, and initialize triplet query  ${\bf{Q}}_t$ through cross-attention with entity visual features ${\bf{Fea}}_e$ to embed visual information and location information. And then, triplet query ${\bf{Q}_t}$ is split into subject query ${\bf{Q}}_s$, predicate query ${\bf{Q}}_p$ and object query ${\bf{Q}}_o$, and input into a self-reasoning decoder for two-by-one self-reasoning training. (3) \textbf{Visual-linguistic Alignment}: The class names of the subject and object, and the triplet they make with the predicate, construct a semantic space with visual-linguistic priors using a CLIP encoder. A visual-linguistic alignment strategy is designed to map the triplet representation to the semantic space to achieve semantic priori auxiliary relational reasoning.

\subsection{Backbone and Entity Decoder}

ResNet50 is adopted as CNN backbone to provide deep visual features for subsequent models, and multi-scale visual feature contexts ${\bf{M}}_e$ is generated by Deformable DETR encoder. The decoder of Deformable DETR is used as the entity decoder to interactively decode with the learnable entity query ${\bf{Q}}_e$. The entity decoder is defined as a mapping function ${{\cal F }_{dec}^{df}}$. The initial entity query ${\bf{Q}}_e$ and the visual feature context ${\bf{M}}_e$ are input to obtain the entity representation ${\bf{H}}_e$ and its associated prediction bounding box ${\bf{B}}_e$,

\begin{equation}\label{eq:1}
{{\bf{H}}_e},{{\bf{B}}_e} = {{\cal F }_{dec}^{df}}\left( {{{\bf{M}}_e},{{\bf{Q}}_e}} \right)
\end{equation}
The bounding box ${\bf{B}}_e$ is mapped back to multi-scale space to obtain the multi-scale entity visual features ${{\bf{Fea}}_e}$,
\begin{equation}\label{eq:2}
{{\bf{Fea}}_e} = Map\left( {{{\bf{M}}_e},{{\bf{B}}_e}} \right)
\end{equation}

\subsection{Self-reasoning Decoder}

\begin{figure*}[tp] \small
	\begin{center}
		\centering
		\epsfig{width=1.5\figurewidth,file=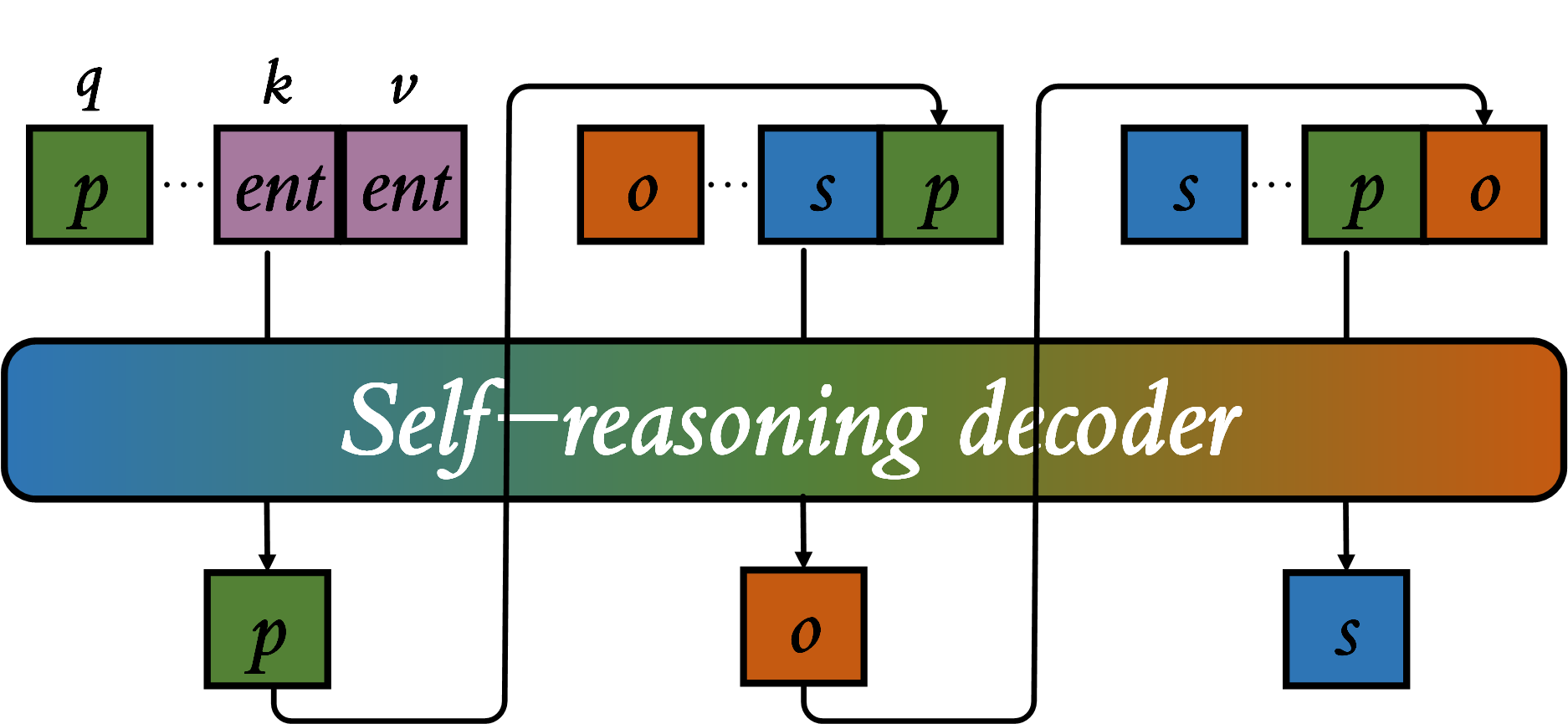}
		\caption{\label{fig:SrDecoder}
			The flowchart of self-reasoning decoder, where \textbf{s+o→p}, \textbf{s+p→o} and \textbf{p+o→s} executed sequentially.}
	\end{center}
	\vspace{-2em}
\end{figure*}

Similar to the entity query in entity decoder, using a fixed set of learnable query is a simple strategy for initializing predicate proposals. However, this triplet query ignores entity candidate visual information and location information. The resulting triplet representation has poor ability to capture structured information and multi-diversity relation reasoning. In order to solve this problem, the multi-scale entity visual features ${{\bf{Fea}}_e}$ and bounding box ${\bf{B}}_e$ are integrated into the process of initializing triplet query. Specifically, the combination of ${{\bf{Fea}}_e}$ and ${\bf{B}}_e$ builds keys and values,
\begin{equation}\label{eq:3}
{\bf{K}}_t^{init} = {\bf{V}}_t^{init} = {\bf{Fe}}{{\bf{a}}_e} + Relu\left( {FC\left( {{{\bf{B}}_e}} \right)} \right)
\end{equation}
where $FC$ denotes the fully connected layer. ${\bf{K}}_t^{init}$ and ${\bf{V}}_t^{init} $ are then encoded with the initialized triplet query ${\bf{Q}}_t^{init}$ using a multi-head cross-attention ($MHA$),
\begin{equation}\label{eq:4}
{{\bf{Q}}_t} = MHA\left( {{\bf{Q}}_t^{init},{\bf{K}}_t^{init},{\bf{V}}_t^{init}} \right)
\end{equation}
Here, the triplet query ${{\bf{Q}}_t}$ has vision and location awareness, and subject query, predicate query and object query are obtained by decoupling predicate query, ${{\bf{Q}}_t} = \{ {{\bf{Q}}_s^{0} },{{\bf{Q}}_p^{0}},{{\bf{Q}}_o^{0}}\}$.  
 
The self-reasoning decoder uses $\{ {{\bf{H}}_e},{{\bf{Q}}_s^{0}},{{\bf{Q}}_p^{0}},{{\bf{Q}}_o^{0}}\}$ and the position encoding $\{ {{\bf{E}}_e},{{\bf{E}}_s},{{\bf{E}}_p},{{\bf{E}}_o}\}$, where ${{\bf{E}}_e}$ is a fixed sine position encoding in Deformable DETR and ${{\bf{E}}_s},{{\bf{E}}_p},{{\bf{E}}_o}$ are the learnable parametric position encoding. A schematic of the self-reasoning decoder is shown in Fig.~\ref{fig:SrDecoder}. The first step of the self-reasoning decoder is to infer the predicate representation ${\bf{H}}_p$, i.e., \textbf{s+o→p}. To incorporate entity representation self-reasoning, we replace subject and object representation with entity representation ${\bf{H}}_e$, and enter ${\bf{H}}_e$ into multi-layer multi-head self-attention via the skip-connected feed-forward network, ${\cal F } \left( {{\bf{Q}},{\bf{K}},{\bf{V}}} \right) = FFN\left( {MHA\left( {{\bf{Q}},{\bf{K}},{\bf{V}}} \right)} \right)$, 
\begin{equation}\label{eq:5}
{\bf{Q}}_p^l = {{\cal F }_{SrD}^{p}} \left( {{\bf{Q}}_p^{l - 1} + {{\bf{E}}_p},{{\bf{H}}_e} + {{\bf{E}}_e},{{\bf{H}}_e} + {{\bf{E}}_e}} \right)
\end{equation}
where the value of $l$ denote as the number of ${{\cal F }_{SrD}^{p}} $ layers, $l={1,2,3}$. The predicate representation ${\bf{H}}_p={\bf{Q}}_p^3$. ${\bf{H}}_p$ and ${\bf{Q}}_s$ are then concatenated and  passed through the FC to form the key of \textbf{s+p}, ${{\bf{K}}_{sp}} = FC\left( {cat\left( {{{\bf{Q}}_s},{{\bf{H}}_p}} \right)} \right)$. After that, ${{\bf{K}}_{sp}}$ is sent to ${{\cal F }_{SrD}^{o}}$ to reason object representation ${\bf{H}}_o$, i.e., \textbf{s+p→o},
\begin{equation}\label{eq:6}
\!\!\!\!\!\!\!\!\!\!\!\!{\bf{Q}}_o^l = {{\cal F }_{SrD}^{o}} \left( {{\bf{Q}}_o^{l - 1} + {{\bf{E}}_o},{{\bf{K}}_{sp}} + {{\bf{E}}_s} + {{\bf{E}}_p},{{\bf{K}}_{sp}} + {{\bf{E}}_s} + {{\bf{E}}_p}} \right)
\end{equation}
The object representation ${\bf{H}}_o={\bf{Q}}_o^3$. Similarly, ${\bf{H}}_o$ and ${\bf{Q}}_p$ are then concatenated and form the key of \textbf{p+o} via FC, ${{\bf{K}}_{po}} = FC\left( {cat\left( {{{\bf{Q}}_p},{{\bf{H}}_o}} \right)} \right)$. The subject representation ${\bf{H}}_s$ obtained by ${{\cal F }_{SrD}^{s}}$,
\begin{equation}\label{eq:7}
{\bf{Q}}_s^l = {{\cal F }_{SrD}^{s}} \left( {{\bf{Q}}_s^{l - 1} + {{\bf{E}}_s},{{\bf{K}}_{po}} + {{\bf{E}}_p} + {{\bf{E}}_o},{{\bf{K}}_{po}} + {{\bf{E}}_p} + {{\bf{E}}_o}} \right)
\end{equation}
The subject representation ${\bf{H}}_s={\bf{Q}}_s^3$. The predicted classes and bounding boxes are obtained by entering ${\bf{H}}_s, {\bf{H}}_p$ and ${\bf{H}}_o$ into the embedding layer. In addition, a visual embedding layer is set up to transform them into 512-dimension to complete the visual-linguistic alignment strategy, i.e., ${\bf{\hat T}} = \{ {{\bf{\hat T}}_s}, {{\bf{\hat T}}_p}, {{\bf{\hat T}}_o} \}$. The embedding layers used here are all fully connected layers.

\subsection{Visual-linguistic Alignment}

Most of one-stage methods only focus on visual modality and ignore the importance of linguistic modality, because the triplet set {\sffamily \textless subject entity, predicate, object entity\textgreater } can be well represented by linguistic features. We introduce a text encoder pre-trained by CLIP to obtain the linguistic features of triplet set, which correspond well to visual features and have visual-linguistic prior knowledge. In the form of ``A photo of $\{\}$'', we select triplet set from the label of each picture and send it to the text encoder to obtain linguistic features ${\bf{ T}} = \{ {{\bf{ T}}_s}, {{\bf{ T}}_p}, {{\bf{ T}}_o} \}$. As shown in Fig.~\ref{fig:Flowchart}, subject corresponds to ``A photo of person'', object corresponds to ``A photo of motorcycle', predicate corresponds to ``A photo of person riding motorcycle'.

To achieve the alignment of visual features and linguistic features of triplet set by class, supervised contrastive learning is performed. Firstly, a supervised contrastive learning is defined as,
\begin{equation}\label{eq:8}
{{\cal L}_{supcon}} =  - \sum\limits_{i = 0}^N {\frac{1}{{|K(i)|}}} \sum\limits_{k \in K(i)} {\log } \frac{{\exp \left( {{{{\bf{x}}_i^T{\bf{x}}_k^ + } \mathord{\left/
					{\vphantom {{{\bf{x}}_i^T{\bf{x}}_k^ + } \tau }} \right.
					\kern-\nulldelimiterspace} \tau }} \right)}}{{\sum\limits_{a \in A(i)} {\exp \left( {{{{\bf{x}}_i^T{\bf{x}}_a^ - } \mathord{\left/
						{\vphantom {{{\bf{x}}_i^T{\bf{x}}_a^ - } \tau }} \right.
						\kern-\nulldelimiterspace} \tau }} \right)} }}
\end{equation}
where for each embedding feature ${{\bf{x}}_i}$ in minibatch, ${K(i)}$ and ${A(i)}$ are the positive and negative sample sets, ${|K(i)|}$ is the number of positive samples, ${{\bf{x}}_k^ + }$ and ${{\bf{x}}_a^ - }$ are one of the positive and negative samples. The predicted triplet is matched to the appropriate ground truth using the Hungarian matching algorithm \cite{HaroldWKuhn1955TheHM}, and the visual features in the prediction and the linguistic features in the ground truth are chosen from ${\bf{t}} = \{ {{\bf{t}}_s}, {{\bf{t}}_p}, {{\bf{ t}}_o} \}$ and ${\bf{\hat t}} = \{ {{\bf{\hat t}}_s}, {{\bf{\hat t}}_p}, {{\bf{\hat t}}_o} \}$ based on the matching index. The alignment losses ${{\cal L}_{vla}}$ of visual to linguistic and linguistic to visual are calculated,

\begin{equation}\label{eq:9}
\begin{array}{l}
{{\cal L}_{vla}} =  - \sum\limits_{i = 0}^N {\frac{1}{{|K(i)|}}} \left( {\sum\limits_{k \in {K_l}(i)} {\log } \frac{{\exp \left( {{{{{{\bf{\hat t}}}^T}{\bf{t}}_k^ + } \mathord{\left/
						{\vphantom {{{{{\bf{\hat t}}}^T}{\bf{t}}_k^ + } \tau }} \right.
						\kern-\nulldelimiterspace} \tau }} \right)}}{{\sum\limits_{a \in {A_l}(i)} {\exp \left( {{{{{{\bf{\hat t}}}^T}{\bf{t}}_a^ - } \mathord{\left/
							{\vphantom {{{{{\bf{\hat t}}}^T}{\bf{t}}_a^ - } \tau }} \right.
							\kern-\nulldelimiterspace} \tau }} \right)} }}} \right.\\
\left. {\begin{array}{*{20}{c}}
	{}&{}
	\end{array} + \sum\limits_{k \in {K_v}(i)} {\log } \frac{{\exp \left( {{{{{\bf{t}}^T}{\bf{\hat t}}_k^ + } \mathord{\left/
						{\vphantom {{{{\bf{t}}^T}{\bf{\hat t}}_k^ + } \tau }} \right.
						\kern-\nulldelimiterspace} \tau }} \right)}}{{\sum\limits_{a \in {A_v}(i)} {\exp \left( {{{{{\bf{t}}^T}{\bf{\hat t}}_a^ - } \mathord{\left/
							{\vphantom {{{{\bf{t}}^T}{\bf{\hat t}}_a^ - } \tau }} \right.
							\kern-\nulldelimiterspace} \tau }} \right)} }}} \right)
\end{array}
\end{equation}
where ${{K_v}(i)}$ and ${{A_v}(i)}$ are the positive and negative sample sets of visual feature, ${{K_l}(i)}$ and ${{A_l}(i)}$ are the positive and negative sample sets of linguistic feature, ${|{K_v}(i)|=|{K_l}(i)|=|K(i)|}$. In addition, the temperature parameter $\tau$, which governs the range of the logits in the softmax, is explicitly optimized as a log-parameterized multiplicative scalar during training. When ${\bf{t}}$ is equal to ${\bf{t}}_p$, ${\bf{t}} = cat\left( {{{\bf{t}}_p},{\bf{t}}_p^{proto}} \right)$, ${\bf{\hat t}} = cat\left( {{{{\bf{\hat t}}}_p},{{{\bf{\hat t}}}_p}} \right)$, where ${\bf{t}}_p^{proto}$ is the prototype of this predicate and reflects the mean of all the features of the predicate of the corresponding index.

%

%
\section{Experimental Results and Discussion}
\label{sec:results}
Experiments using the Visual Genome dataset are conducted to validate the proposed SrTR. For comparison algorithms, several state-of-the-art SGG algorithms are used, including two-stage algorithms, RelDN \cite{ZhangJi2019GraphicalCL}, VCTree-TDE \cite{KaihuaTang2020UnbiasedSG}, G-RCNN \cite{JianweiYang2018GraphRF}, Motifs \cite{RowanZellers2017NeuralMS}, GPS-Net \cite{XinLin2020GPSNetGP}, KERN \cite{TianshuiChen2019KnowledgeEmbeddedRN}, BGNN \cite{RongjieLi2021BipartiteGN}, IMP \cite{DanfeiXu2017SceneGG}, CISC \cite{WenbinWang2019ExploringCA}, one-stage algorithms, FCSGG\cite{HengyueLiu2021FullyCS}, RelTR \cite{YurenCong2022RelTRRT}. The R@K and mR@K are employed to evaluate performance.

\subsection{Experimental Setting}
Visual Genome contains a total of 108k images with 150 entity classes and 50 predicate classes. 70\% of the images are used as training set, and the remaining 30\% are used as test set. 5k images are extracted from the training set for validation. There are three evaluation settings: (1) Predicate classification (PredCLS/PredDET): Given the bounding box and object label, the model predicts possible predicates between objects; (2) Scene graph classification (SGCLS): Given the bounding box, the model predicts the object label and the predicate relation between objects; (3) SGDET: Bounding box, object label, and predicate between objects are directly predicted by the model.

SrTR is implemented on the Pytorch platform. All the experiments are conducted on 8 Nvidia GTX 3090 GPU. The initial learning rates of ResNet50 backbone and  Transformer are set to 10-5 and 10-4, respectively. The number of encoder and decoder layers in SrTR is set to 3, and multi-head attention modules with 8 heads are used. The number of entity query and triplet query are set to 100 and 200.

\begin{table}[]
	\caption{\label{table:Ablation}
		Self-reasoning Decoder (SrD) and the Visual-linguistic Alignment (VLA) are isolated separately from the framework.}
	\begin{center}
		\begin{tabular}{cc|cccc}
			\hline \hline
			\multicolumn{2}{c|}{Ablation Setting} & \multicolumn{4}{c}{SGDET}                                   \\
			SrD               & VLA               & R@20          & R@50          & mR@20        & mR@50        \\ \hline
			×                 & ×                 & 18.7          & 21.9          & 4.5          & 6.3          \\
			$\surd$                & ×                 & 20.1          & 23.8          & 5.6          & 7.7          \\
			×                 & $\surd$                 & 19.5          & 22.3          & 5.0          & 6.9          \\
			$\surd$                 & $\surd$                 & \textbf{20.5} & \textbf{24.7} & \textbf{6.1} & \textbf{8.4} \\ \hline \hline
		\end{tabular}
	\end{center}
\end{table}
\subsection{Ablation Study}

The self-reasoning decoder (SrD) is the key components of SrTR, and visual-linguistic alignment (VLA) is the main strategy for introducing visual-linguistic knowledge. Ablation analyses are carried out by eliminating each component from the total framework in order to evaluate the contribution of important SrTR components. It is clear from Table \ref{table:Ablation} that both SrD and VLA improve the performance of baseline when using one of them, and SrD is more promising, with an increase of about 1.2 in mR@. When SrD and VLA are simultaneously added to SrTR, the mR@ increased by about 1.8.

\begin{table*}[]
	\vspace{-0.5em}
	\setlength{\tabcolsep}{0.1em}
	\caption{\label{table:VG}
		Comparison with state-of-the-art scene graph generation methods on Visual Genome test set. These methods are divide into two-stage and one-stage. The best numbers in two-stage methods are shown in blue bold, and the best numbers in one-stage methods are shown in red bold.}
	\begin{center}
		\begin{tabular}{cc|c|cccc|cccc|cccc|c}
			\hline \hline
			\multicolumn{2}{c|}{\multirow{2}{*}{Method}}                                                           & \multirow{2}{*}{$AP_{50}$} & \multicolumn{4}{c|}{PredCLS}                                  & \multicolumn{4}{c|}{SGCLS}                                    & \multicolumn{4}{c|}{SGDET}                                   & \multirow{2}{*}{\#params(M)} \\
			\multicolumn{2}{c|}{}                                                                                  &                            & R@20          & R@50          & mR@20         & mR@50         & R@20          & R@50          & mR@20         & mR@50         & R@20          & R@50          & mR@20        & mR@50         &                              \\ \hline
			\multicolumn{1}{c|}{\multirow{9}{*}{\begin{tabular}[c]{@{}c@{}}two\\ stage\end{tabular}}} & MOTIFS     & 20.0                       & 58.5          & 65.2          & 10.8          & 14.0          & 32.9          & 35.8          & 6.3           & 7.7           & 21.4          & 27.2          & 4.2          & 5.7           & 240.7                        \\
			\multicolumn{1}{c|}{}                                                                     & KERN       & 20.0                       & 59.1          & 65.8          & -             & 17.7          & 32.2          & 36.7          & -             & 9.4           & 22.3          & 27.1          & -            & 6.4           & 405.2                        \\
			\multicolumn{1}{c|}{}                                                                     & RelDN      & -                          & 66.9          & 68.4          & -             & -             & 36.1          & 36.8          & -             & -             & 21.1          & 28.3          & -            & -             & 615.6                        \\
			\multicolumn{1}{c|}{}                                                                     & VCTree-TDE & 28.1                       & 39.1          & 49.9          & 17.2          & 23.3          & 22.8          & 28.8          & 8.9           & 11.8          & 14.3          & 19.6          & 6.3          & 9.3           & 360.8                        \\
			\multicolumn{1}{c|}{}                                                                     & GPS-Net    & -                          & \color{blue}{\textbf{67.6}} & \color{blue}{\textbf{69.7}} & \color{blue}{\textbf{17.4}} & 21.3          & \color{blue}{\textbf{41.8}} & \color{blue}{\textbf{42.3}} & \color{blue}{\textbf{10.0}} & 11.8          & 22.3          & 28.9          & 6.9          & 8.7           & -                            \\
			\multicolumn{1}{c|}{}                                                                     & BGNN       & 29.0                       & -             & 59.2          & -             & \color{blue}{\textbf{30.4}} & -             & 37.4          & -             & \color{blue}{\textbf{14.3}} & \color{blue}{\textbf{23.3}} & \color{blue}{\textbf{31.0}} &\color{blue}{\textbf{7.5}} & \color{blue}{\textbf{10.7}} & 341.9                        \\
			\multicolumn{1}{c|}{}                                                                     & IMP        & -                          & 58.5          & 65.2          & -             & 9.8           & 31.7          & 34.6          & -             & 5.8           & 14.6          & 20.7          & -            & 3.8           & \color{blue}{\textbf{203.8}}              \\
			\multicolumn{1}{c|}{}                                                                     & CISC       & -                          & 42.1          & 53.2          & -             & -             & 23.3          & 27.8          & -             & -             & 7.7           & 11.4          & -            & -             & -                            \\
			\multicolumn{1}{c|}{}                                                                     & G-RCNN     & 24.8                       & -             & 54.2          & -             & -             & -             & 29.6          & -             & -             & -             & 11.4          & -            & -             & -                            \\ \hline
			\multicolumn{1}{c|}{\multirow{3}{*}{\begin{tabular}[c]{@{}c@{}}one\\ stage\end{tabular}}} & FCSGG      & 28.5                       & 33.4          & 41.0          & 4.9           & 6.3           & 19.0          & 23.5          & 2.9           & 3.7           & 16.1          & 21.3          & 2.7          & 3.6           & 87.1                         \\
			\multicolumn{1}{c|}{}                                                                     & RelTR      & 26.4                       & 63.1          & 64.2          & 19.5          & 20.7          & \color{red}{\textbf{28.8}} & \color{red}{\textbf{36.3}} & 7.5           & 10.6          & 20.1          & \color{red}{\textbf{24.8}} & 5.6          & 7.7           & 63.7                         \\
			\multicolumn{1}{c|}{}                                                                     & SrTR       & 23.1                       & \color{red}{\textbf{64.8}} & \color{red}{\textbf{65.7}} & \color{red}{\textbf{21.1}} & \color{red}{\textbf{22.3}} & 27.0          & 35.6          & \color{red}{\textbf{8.9}}  & \color{red}{\textbf{12.1}} & \color{red}{\textbf{20.5}} & 24.7          & \color{red}{\textbf{6.1}} & \color{red}{\textbf{8.4}}  & \color{red}{\textbf{53.4}}                         \\ \hline \hline
		\end{tabular}
	\end{center}
\end{table*}

\subsection{Performance on Visual Genome}
\label{sec:Classification Performance}

\begin{figure*}[tp] \small
	\begin{center}
		\centering
		\epsfig{width=2\figurewidth,file=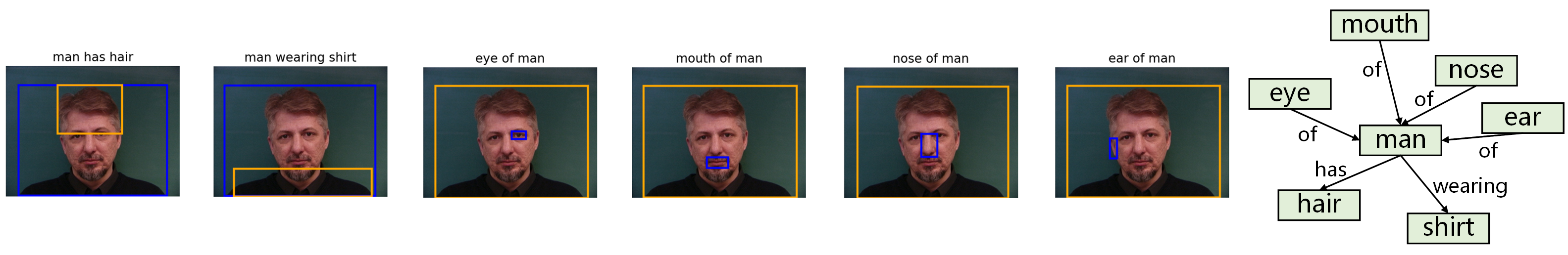}
		\caption{\label{fig:sg}
			Qualitative results for scene graph generation of Visual Genome dataset.}
	\end{center}
	\vspace{-2em}
\end{figure*}

\begin{figure*}[tp] \small
	\begin{center}
		\centering
		\epsfig{width=2\figurewidth,file=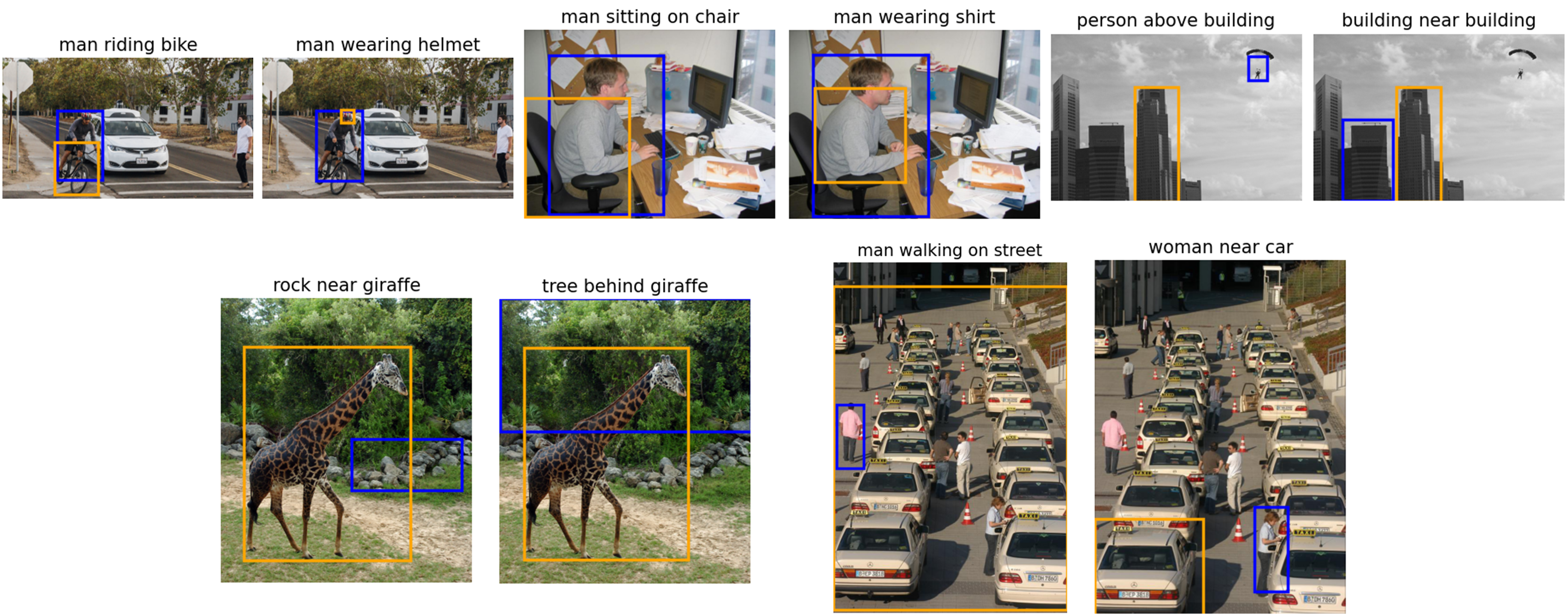}
		\caption{\label{fig:vis}
			Qualitative results of examples of rational relational inference. The blue box is the subject box and the orange box is the object box.}
	\end{center}
	\vspace{-2em}
\end{figure*}

The results for R@K and mR@K are given in Table \ref{table:VG} in the two-stage and one-stage methods, the best recall is represented in blue bold for the two-stage method and in red bold for the one-stage method. The mR@K metric of the proposed SrTR in three evaluation settings is higher than that of the one-stage model FCSGG and RelTR, and the number of parameters is also the lowest. On PredCLS, SrTR reached mR@20=21.1, mR@50=22.3, 1.6 higher than RelTR. Since the performance of the entity detector in SrTR is only 23.1, its advantages in SGCLS and SGDET are not obvious, but there is a slight improvement on mR@20/mR@50. Compared with the one-stage model, the two-stage model has better performance. The entity detection accuracy of BGNN is as high as 29.0, which is about 2 higher than SrTR on SGDET. However, they have many parameters, and even the IMP with the lowest parameters is 3.8 times that of SrTR, which is not conducive to lightweight deployment in practical applications. In general, SrTR has strong competitiveness and can better capture the relationship between entities when there are entity classes and bounding boxes and only the relationship between entities is considered.

Qualitative results for scene graph generation of Visual Genome dataset are illustrated in Fig. \ref{fig:sg}, where the blue box is the subject box and the orange box is the object box. We show 6 relationships with the highest confidence scores and the generated scene graphs. There is a significant deviation in the training data set in Visual Genome. As the most frequent predicate, ``on'' is the most easily predicted relationship, and the probability of ``of'' being mispredicted as ``on'' is the highest. As shown in Fig. \ref{fig:sg}, in most cases, ``eye of man'' is more appropriate than ``eye on man''. This indicates that SrTR predicts predicates that are more consistent with the subject and object semantic relationships. In addition, Qualitative results of examples of rational relational inference are shown in Fig. \ref{fig:vis}, which are selected from the top 10 relationships with the highest confidence scores. Among them, sitting on, above, near and walking on are low-frequency predicates in the training set, which can be inferred in SrTR with higher confidence scores.

\section{Conclusions}
\label{sec:conclusions}
From the perspective of self-reasoning, a Self-reasoning Transformer with Visual-linguistic Knowledge (SrTR) is proposed, which enhances the model to relation reasoning capacity. A self-reasoning decoder is designed to realize the reasoning between elements in { \sffamily \textless subject entity, predicate, object entity \textgreater }. Furthermore, considering the importance of linguistic modality in representing the relationship between entities, we introduce the visual-linguistic prior information of the large-scale pre-training image-text foundation model CLIP, and design visual-linguistic alignment as triplet representation to embed prior knowledge in the training process. Experiments on the Visual Genome dataset show that SrTR has competitive performance compared to other state-of-the-art one-stage methods.

\bibliographystyle{IEEEtran}
\bibliography{bibfile_zyx}

\end{document}